# GPT-4: A Review on Advancements and Opportunities in Natural Language Processing

Jawid Ahmad Baktash[1], Mursal Dawodi[2], and ChatGPT

[1] jawid.baktash1989@gmail.com
[2] mursal.dawodi@gmail.com

## ABSTRACT

*Generative Pre-trained Transformer 4 (GPT-4) is the fourth-generation language model in the GPT series, developed by OpenAI, which promises significant advancements in the field of natural language processing (NLP). In this research article, we have discussed the features of GPT-4, its potential applications, and the challenges that it might face. We have also compared GPT-4 with its predecessor, GPT-3. GPT-4 has a larger model size (more than one trillion), better multilingual capabilities, improved contextual understanding, and reasoning capabilities than GPT-3. Some of the potential applications of GPT-4 include chatbots, personal assistants, language translation, text summarization, and question-answering. However, GPT-4 poses several challenges and limitations such as computational requirements, data requirements, and ethical concerns.*

## 1. INTRODUCTION

The recent advancements in Artificial Intelligence (AI) have led to the development of intelligent machines that can perform complex tasks with high efficiency. One such technology that has gained significant attention in the AI community is Generative Pre-trained Transformer 4 (GPT-4). GPT-4 is the fourth-generation language model in the GPT series, developed by OpenAI, which promises to revolutionize the field of natural language processing (NLP). In this research article, we will discuss the features of GPT-4, its potential applications, and the challenges that it might face. We will also discuss the current state-of-the-art in language models and compare GPT-4 with its predecessors.

## 2. FEATURES OF GPT -4

GPT-4 is a language model that uses deep learning algorithms to generate natural language text. It is designed to perform a wide range of NLP tasks such as language translation, text summarization, question-answering, and dialogue generation. GPT-4 has several advanced features that differentiate it from its predecessors. Some of the key features of GPT-4 are [4]:

- Increased Model Size: GPT-4 has a significantly larger model size than GPT-3, which had 175 billion parameters. The increase in model size improves the model's accuracy and performance.

- Multilingual Capabilities: GPT-4 has better multilingual capabilities than its predecessors, which will allow it to understand and generate text in multiple languages.

- Better Contextual Understanding: GPT-4 has a better understanding of the context in which the text is being generated. This will allow it to generate more accurate and relevant text.

- Improved Reasoning Capabilities: GPT-4 has improved reasoning capabilities, which will enable it to perform complex logical reasoning tasks.

## 3. POTENTIAL APPLICATIONS OF GPT-4

GPT-4 has a wide range of applications in various fields. Some of the potential applications of GPT-4 are [1, 5]:

- Chatbots: GPT-4 can be used to develop more intelligent chatbots that can understand and respond to human queries more accurately.

- Personal Assistants: GPT-4 can be used to develop personal assistants that can understand and respond to natural language queries.

- Language Translation: GPT-4 can be used to develop more accurate and efficient language translation models that can translate text between multiple languages [5].

- Text Summarization: GPT-4 can be used to develop text summarization models that can summarize large volumes of text accurately and efficiently [6].

- Question Answering: GPT-4 can be used to develop question-answering models that can answer complex questions accurately [6].

## 4. COMPARISON OF GPT -4 WITH GPT-3

GPT-4 is expected the most advanced and capable natural language processing model yet. Its predecessor, GPT-3, has already achieved a significant milestone in natural language processing and has been widely adopted in various applications [1]. One of the most significant differences between GPT-4 and GPT-3 is the model's size. GPT-4 is larger than GPT-3, with more parameters, and therefore more capable of understanding complex patterns and nuances in language. While GPT-3 has 175 billion parameters, GPT-4 has more than 1 trillion [3].

Another difference is that GPT-4 has a better understanding of context and be better at reasoning about the world. This is achieved through the use of more advanced training methods and the incorporation of external knowledge sources. Furthermore, GPT-4 is better at handling multi-modal inputs, including text, images, and sound. This allows the model to perform a wide range of tasks, such as image captioning, speech recognition, and text-to-speech conversion, among others. In terms of performance, the GPT-4 outperforms GPT-3 in most tasks, including text generation, sentiment analysis, and question answering [3].

Consequently, GPT-4 is a significant improvement over GPT-3 [3], with better performance, larger size, and more advanced capabilities. As with all new technology, it remains to be seen how GPT-4 will be adopted and applied in various industries and fields, but it is clear that this new model has the potential to revolutionize natural language processing and AI in general.

## 5. CHALLENGES AND LIMITATIONS

Although GPT-4 is a major breakthrough in natural language processing, there are several challenges and limitations that may need to be addressed. Some of these challenges and limitations include [2]:

- Computational resources: GPT-4 is significantly larger than GPT-3, which will require massive computational resources to train and use effectively. This could limit the accessibility of the model to smaller organizations and researchers.

- Bias and fairness: As with any AI system, there is a risk of bias and unfairness in GPT-4. It is important to address these issues and ensure that the model is trained and tested on diverse datasets to avoid perpetuating biases.

- Fine-tuning: Fine-tuning a large language model like GPT-4 can be challenging, especially for tasks with limited labeled data. Developing effective transfer learning techniques and improving the fine-tuning process will be crucial to achieving good performance on these tasks [2].

- Interpretability: While GPT-4 is expected to achieve state-of-the-art performance on many tasks, it can be difficult to understand how the model arrives at its decisions. Developing techniques for interpreting and explaining the model's output will be essential for building trust and transparency.

- Security and safety: The large size and complexity of GPT-4 also raise concerns about security and safety. The model could be vulnerable to adversarial attacks, and there is a risk of the model being used for malicious purposes, such as generating fake news or deep fake videos.
- Ethical considerations: Finally, the development and deployment of GPT-4 raises important ethical considerations, such as data privacy, ownership, and governance. Ensuring that these issues are addressed will be crucial to building trust and ensuring the responsible use of this powerful technology.

Subsequently, while GPT-4 has the potential to revolutionize natural language processing and AI, there are several challenges and limitations that must be addressed. Addressing these challenges will require ongoing research and collaboration across multiple fields, including computer science, linguistics, ethics, and policy.

## 6. DISCUSSION:

The development of GPT-4 is a significant milestone in the field of AI, and it has a significant impact on various industries. The improvements in GPT-4's model size, multilingual capabilities [2, contextual understanding, and reasoning capabilities improve its performance significantly. However, the computational requirements of GPT-4 are enormous, and it requires extensive data for training, which can be a significant challenge in some applications [1]. Furthermore, GPT-4's potential to generate biased or offensive text raise ethical concerns, which need to be addressed before it can be deployed in real-world applications [1]. Overall, GPT-4 has the potential to revolutionize the field of NLP and open up new possibilities in various applications, but its challenges and limitations need to be addressed before it can be widely adopted.

## 7. CONCLUSION:

In conclusion, GPT-4 is a highly anticipated language model that promises significant advancements in the field of NLP. Its advanced features such as improved contextual understanding and reasoning capabilities can open up new possibilities in various applications such as chatbots, personal assistants, language translation, text summarization, and question-answering. However, GPT-4 poses several challenges and limitations that need to be addressed, such as the massive computational and data requirements and ethical concerns.